\documentclass{article}
\usepackage{arxiv}

\usepackage[utf8]{inputenc} 
\usepackage[T1]{fontenc}    
\usepackage{hyperref}       
\usepackage{url}            
\usepackage{booktabs}       
\usepackage{nicefrac}       
\usepackage{microtype}      
\usepackage{lipsum}
\usepackage{graphicx}

\newcommand{\mat}[1]{\mathbf{#1}}

\usepackage{amssymb,amsfonts}
\usepackage{subfig}
\usepackage{dblfloatfix}
\usepackage{mwe}

\usepackage{times}
\usepackage{soul}
\usepackage{booktabs}
\usepackage{scalerel}
\usepackage{amsmath}

\title{Attend And Discriminate: Beyond the State-of-the-Art for Human Activity Recognition using Wearable Sensors}

\author{
Alireza Abedin \\
School of Computer Science\\
The University of Adelaide\\
alireza.abedinvaramin@adelaide.edu.au \\
  \And
 Mahsa Ehsanpour \\
School of Computer Science\\
The University of Adelaide\\
mahsa.ehsanpour@adelaide.edu.au \\
 \And
 Qinfeng Shi\\
School of Computer Science\\
The University of Adelaide\\
javen.shi@adelaide.edu.au \\
  \And
  Hamid Rezatofighi \\
School of Computer Science\\
The University of Adelaide\\
hamid.rezatofighi@adelaide.edu.au \\
   \And
Damith C. Ranasinghe \\
School of Computer Science\\
The University of Adelaide\\
damith.ranasinghe@adelaide.edu.au \\
}

\begin{document}
\maketitle

\begin{abstract}
  Wearables are fundamental to improving our understanding of human activities, especially for an increasing number of healthcare applications from rehabilitation to fine-grained gait analysis. Although our collective know-how to solve Human Activity Recognition (HAR) problems with wearables has progressed immensely with end-to-end deep learning paradigms, several fundamental opportunities remain overlooked. We rigorously explore these new opportunities to learn enriched and highly discriminating activity representations. We propose: i)~learning to exploit the \textit{latent} relationships between multi-channel sensor modalities and specific activities; ii)~investigating the effectiveness of \textit{data-agnostic augmentation} for multi-modal sensor data streams to regularize deep HAR models; and iii)~incorporating a classification loss criterion to encourage minimal intra-class representation differences whilst maximising inter-class differences to achieve more discriminative features. Our contributions achieves \textit{new} state-of-the-art performance on \textit{four} diverse activity recognition problem benchmarks with large margins---with up to~6\% relative margin improvement. We extensively validate the contributions from our design concepts through extensive experiments, including \textit{activity misalignment} measures, \textit{ablation} studies and insights shared through both quantitative and qualitative studies. 
\end{abstract}

\keywords{activity recognition; deep learning; attention; wearable sensors; time-series data}


\section{Introduction}\label{sa_sec:intro}

Wearable sensors provide an infrastructure-less multi-modal sensing method. Current trends point to a pervasive integration into our lives with wearables providing the basis for wellness and healthcare applications from rehabilitation, caring for a growing older population to improving human performance~\cite{sparsesense,mannini2017activity,Govercin2010UserReqFallDetect,healthapp,pd,roberto2018ponealarms,health1,health2,health3,health4,health5}. Fundamental to these applications is our ability to automatically and accurately recognize human activities from often tiny sensors embedded in wearables. 

Wearables capture individuals' activity dynamics by continuously recording measurements through different sensor channels over time and generate \textit{multi-channel time-series} data streams. Consequently, the problem of human activity recognition (HAR) with wearables involves temporal localization and classification of actions embedded in the generated sequences. Adoption of deep neural networks for HAR has created pipelines for end-to-end learning of activity recognition models yielding state-of-the-art (SoA) performance \cite{deepconvlstm,bilstm,fcn,attention}.

\subsection{Problem and Motivations}
Despite progress towards end-to-end deep learning architectures for achieving state-of-the-art performance on HAR problems, we uncover key under explored dimensions with significant potential for improving the performance of the state-of-the-art frameworks: 
\begin{itemize}
	\item HAR data acquisition often involves recording of motion measurements over number of sensors and channels. Therefore, we can expect the capability of different sensor modalities and channels to capture and encode some activities better than others whilst having complex interactions between sensors, channels and activities. Thus, we hypothesize that learning to exploit the relationships between multi-channel sensor modalities and specific activities can contribute to learning enriched activity representations---\textit{this insight remains unstudied}.
	
	\item Human actions, for example walking and walking up-stairs, exhibit significant intra-class variability and inter-class similarities. This suggests imposing optimization objectives for training that not only ensure class separability but also encourage compactness in the established feature space. However, \textit{the commonly adopted cross-entropy loss function does not jointly accommodate both objectives}. 

	\item Due to the laborious process of collecting annotated sequences with wearables, sensor HAR datasets are often small in size. While expanding training data with virtual samples has proved beneficial in achieving better generalization performance for computer vision tasks, \textit{data-agnostic augmentation of multi-modal sensor data streams remains less explored, while data augmentation in general remains under-utilized for HAR}.

\end{itemize}

\subsection{Our Contributions}

\begin{figure*}[tb]
	\centering
	\includegraphics[width=\linewidth]{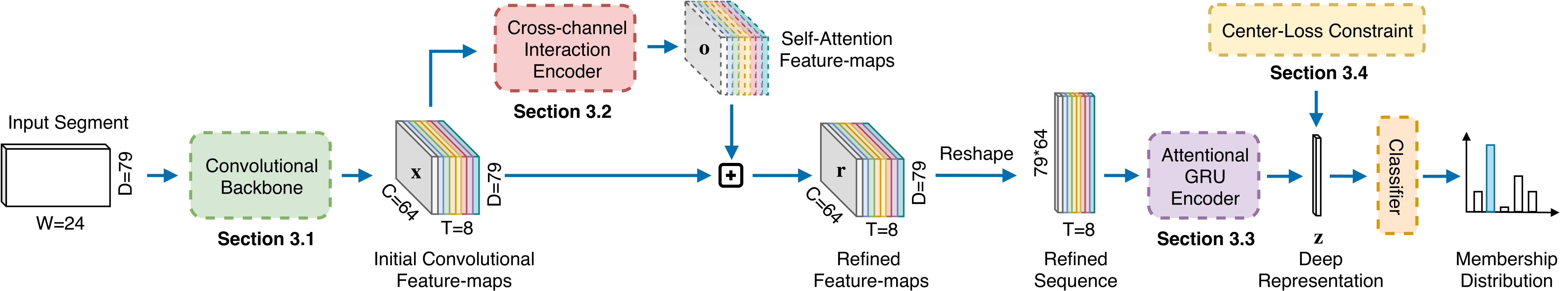}
	\caption{An overview of our proposed HAR framework}
	\label{sa_fig:arch}
\end{figure*}

Motivated by the opportunities to further HAR research, our \textit{key contribution} is to propose a \textit{new HAR framework} built upon multiple architectural elements and demonstrate its capability to realize new state-of-the-art performance that generalizes across multiple diverse wearable sensor datasets. We illustrate our framework in Fig.~\ref{sa_fig:arch} and summarize our key contributions below:

\begin{enumerate}

	\item  We propose and design a \textit{cross-channel interaction encoder} to incorporate a self-attention mechanism to learn to exploit the different capabilities of sensor  modalities and latent interactions between multiple sensor channels capturing and encoding activities. The encoder module captures latent correlations between multi-sensor channels to generate self-attention feature maps to enrich the convolutional feature representations (\textbf{Section~\ref{sa_sec:cie}} and \textbf{Fig.~\ref{sa_fig:sa}}). Subsequently, we enhance this sequence of enriched features by an \textit{attentional GRU encoder} to capture the relevant temporal context (\textbf{Section~\ref{sa_sec:age}}).

	\item In recognizing the intra-class variations of HAR activities, we incorporate the \textit{center-loss criterion} into our framework to encourage minimal intra-class representation differences whilst maximising inter-class differences to achieve more discriminative features (\textbf{Section~\ref{sa_sec:centerloss}}).

	\item In recognizing the difficulty of multi-modal sensor data augmentation and the general under-utility of data augmentation methods in wearable HAR problems; we investigate and show the effectiveness and seamless integration of the recent data-agnostic \textit{mixup} method for multi-modal sensor data augmentation (\textbf{Section~\ref{sa_sec:mixup}}).

	\item Under a unified evaluation protocol, our proposed framework achieves significant improvements against the state-of-the-art on four diverse HAR benchmarks and, thus, highlights  the effectiveness and generalizability of our framework (\textbf{Section~\ref{sa_sec:exp_comp}}). Further, we share our insights through extensive quantitative and qualitative results as well as an ablation study to comprehensively demonstrate the contributions made by the architectural elements in our new HAR framework (\textbf{Section~\ref{sa_sec:ablation}}).

\end{enumerate} 
\section{Related Work}

Traditionally, the standard activity recognition pipeline for time-series sensory data involved sliding window segmentation, manual hand-crafted feature design, and subsequent activity classification with classical machine learning algorithms \cite{bulling}. Studies along these line have extensively explored statistical~\cite{Bao:2004,Ravi:2005}, basis transform~\cite{Huynh:2005}, multi-level~\cite{Zhang:2012}, and bio-mechanical features~\cite{wickramasinghe2017sequence} coupled with the employment of shallow classifiers including support vector machines \cite{Bulling:2012}, decision trees \cite{Bao:2004}, joint boosting \cite{Lara:2012}, graphical models~\cite{ShinmotoTorres:2017CRF}, and multi-layer perceptrons \cite{Randell:2000}. While this manually tuned procedure has successfully acquired satisfying results for relatively simple recognition tasks, its generalization performance is limited by heavy reliance on domain expert knowledge to engineer effective features.

Over the past years, the emerging paradigm of deep learning has presented unparalleled performance in various research areas including computer vision, natural language processing and speech recognition \cite{Lecun:2015}. When applied to sensor-based HAR, deep learning allows for automated end-to-end feature extraction and thus, largely alleviates the need for laborious feature engineering procedures. Motivated by these, sensor-based human activity recognition has witnessed extensive and successful adoption of deep learning paradigms in diverse HAR application settings \cite{nabil,misra,roy}. 

Pioneering studies in the field have explored Restricted Boltzmann Machines (RBMs) for automatic representation learning \cite{rbm1,alsheikh,pd,rbm2}. Recently, deep architectures based on convolutional neural networks (CNNs) have been predominantly leveraged to automate feature extraction from sensor data streams while mutually enhancing activity classification performance \cite{cnnmobile,multichannelcnn,cnnphone,cnnnew}. These studies typically employ a cascaded hierarchy of 1D convolution filters along the temporal dimensions to capture salient activity features at progressively more abstract resolutions. The acquired latent features are ultimately unified and mapped into activity class scores using a fully connected network. Taking a different approach, \cite{fcn} presents an efficient dense labeling architecture based on fully convolutional networks that allows making activity predictions for every sample of a sliding window segment. 

Another popular architecture design for HAR adopts deep recurrent neural networks (RNNs) that leverage memory cells to directly model temporal dependencies between subsequent sensor samples. In particular, \cite{bilstm} investigates forward and bi-directional long short-term memory (LSTM) networks, and \cite{ensemble} explores ensemble of diverse LSTM learners to exploit the sequential nature of sensor data. Combining these concepts, \cite{deepconvlstm} proposes DeepConvLSTM by pairing convolutional and recurrent networks in order to model the temporal correlations at a more abstract representation level. In \cite{attention}, the recurrent network of DeepConvLSTM is expanded with attention layers to model the relevant temporal context of sensor data. We refer readers to \cite{survey} for a curated list of recent HAR studies with deep neural networks.

Despite the great progress in the field, we can see that the unique opportunities we discussed in Section~\ref{sa_sec:intro}  for learning from multi-channel time-series data generated by body-worn sensors remain. Conventionally, the feature-maps generated by convolutional layers are trivially vectorized and fed to fully connected layers or recurrent networks to ultimately produce classification outcomes. However, such manipulation of the convolutional feature-maps fails to explicitly capture and encode the inter-channel interactions that can aid accurate recognition of activities. Moreover, regardless of the architectural designs, cross-entropy loss constitutes the common choice for supervised training of deep HAR models. Yet, this optimization objective alone does not cater for the need to achieve minimal intra-class compactness of feature representations \cite{centerloss} necessary to counter the significant intra-class variability of human activities. In addition, while data augmentation has shown great potential for regularizing deep neural networks in the computer vision domain, the effectiveness of data augmentation for multi-channel time-series data captured by wearables remains under-utilized for HAR. 
\section{Our Proposed HAR Framework}\label{sa_sec:method}

The goal is to develop an end-to-end deep HAR model that directly consumes raw sensory data captured by wearables and seamlessly outputs precise activity classification decisions. In our framework, a network composed of 1D convolutional layers serves as the backbone feature extractor in order to automatically extract an initial feature representation for each sensory segment. Subsequently, we propose a two-staged refinement process to enrich the initial feature representations prior to classification that allows the model to $i$) effectively uncover and encode the underlying sensor channel interactions at each time-step, and $ii$) learn the relevant temporal context within the sequence of refined representations. Moreover, we encourage intra-class compactness of representations with center-loss while regularizing the network with mixup data augmentation during training. In what follows, we elaborate on the components of our framework, illustrated in Fig. \ref{sa_fig:arch}.

\subsection{1D Convolutional Backbone} 
Following the sliding window segmentation, the input to the network is a slice of the captured time-series data $x\in\mathbb{R}^{\textrm{D}\times \textrm{W}}$, where $\textrm{D}$ denotes the number of sensor channels used for data acquisition and $\textrm{W}$ represents the choice for the window duration. For automatic feature extraction, the input is then processed by a convolutional backbone operating along the temporal dimension. Given the 1D structure of the adopted filters, progressively more abstract temporal representations are learned from nearby samples without fusing features in-between different sensor channels. Ultimately, the backbone yields a feature representation $\mat{x}\in\mathbb{R}^{\textrm{C}\times \textrm{D} \times \textrm{T}}$, where in each of the $\textrm{C}$ feature maps, the sensor channel dimension $\textrm{D}$ is preserved while the temporal resolution is down-sampled to $\textrm{T}$. Without loss of generality, in this paper we employ the convolutional layers of a state-of-the-art HAR model \cite{deepconvlstm} as the backbone feature extractor; the input segment is successively processed by four layers, each utilizing 64 one-dimensional filters of size 5 along the temporal axis with ReLU non-linearities. 

\subsection{Cross-Channel Interaction Encoder (CIE)}\label{sa_sec:cie}

Accurate realization of fine-grained human actions using wearables is often associated with utilizing multitude of on-body sensing devices that capture activity data across multiple channels. Measurements captured by different sensor channels provide different views of the same undergoing activity and are thus, inherently binded together in an unobservable latent space. Accordingly, we seek to design an end-to-end trainable module that takes as input the initial convolutional feature-maps at each time-step, learns the interactions between any two sensor channels within the feature-maps, and leverages this overlooked source of information to enrich the sensory feature representations for HAR. 

Motivated by the emerging successful applications of self-attention \cite{attentionis,nonlocal,selfgan} in capturing global dependencies by computing relations at any two positions of the input, here we design a \textit{Cross-Channel Interaction Encoder (CIE)} that adopts self-attention mechanism to effectively process the initial feature representations and uncover the latent channel interactions. To this end, we first compute the normalized correlations across all pairs of sensor channel features $\mat{x}_{t}^{d}$ and $\mat{x}_{t}^{d^\prime}$ using the embedded Gaussian function at each time-step $t$,

\begin{equation}\label{sa_eq:att}
\mat{a}_{t}^{d,d^{\prime}} = \frac{\exp\Big(f(\mat{x}_{t}^{d})^{\intercal}g(\mat{x}_{t}^{d^\prime})\Big)}{\sum_{d^\prime=1}^{\textrm{D}}\exp\Big(f(\mat{x}_{t}^{d})^{\intercal}g(\mat{x}_{t}^{d^{\prime}})\Big)},
\end{equation}
where $\mat{a}_{t}^{d,d^{\prime}}$ indicates the attendance of our model to the features of sensor channel $d^\prime$ when refining representations for sensor channel $d$. Subsequently, the extracted correlations are leveraged in order to compute the response for the $d^{\textrm{th}}$ sensor channel features $\mat{x}_t^d\in\mathbb{R}^{\textrm{C}}$ and generate the corresponding self-attention feature-maps $\mat{o}_{t}^{d}$ at each time-step

\begin{equation}
\mat{o}_{t}^{d} = v\Big(\sum_{d^\prime=1}^{\textrm{D}} \mat{a}_{t}^{d,d^{\prime}} h(\mat{x}_{t}^{d^\prime})\Big).
\end{equation}

Technically, the self-attention in the CIE module functions as a non-local operation which computes the response for sensor channel $d$ at each time-step by attending to all present sensor channels' representations in the feature-maps at the same time-step. In the above, $f$, $g$, $h$, and $v$ all represent linear embeddings with learnable weight matrices ($\in\mathbb{R}^{\textrm{C}\times\textrm{C}}$) that project feature representations into new embedding spaces where computations are carried out. Having obtained the self-attention feature-maps, the initial feature-maps are then added back via a residual link (indicated by $\bigoplus$ in Fig. \ref{sa_fig:arch}) to encode the interactions and generate the refined feature representations $\mat{r}_t$,  

\begin{equation}
\mat{r}_t^{d} = \mat{o}_{t}^{d}+\mat{x}_t^{d}. 
\end{equation}
With the residual connection in place, the model can flexibly decide to use or discard the correlation information. During training, the HAR model leverages the CIE module to capture the interactions between different sensor channels. The discovered correlations are encoded inside the self-attention weights and leveraged at inference time to help support the model's predictions. 

\subsection{Attentional GRU Encoder (AGE)} \label{sa_sec:age} 

As a result of employing the CIE module, the feature-maps generated at each time-step are now contextualized with the underlying cross-channel interactions. As shown in Fig. \ref{sa_fig:arch}, we vectorize these representations at each time-step to obtain a sequence of refined feature vectors $(\mat{r}_t\in\mathbb{R}^{\textrm{CD}})_{t=1}^{\textrm{T}}$ ready for sequence modeling. Given that not all time-steps equally contribute in recognition of the undergoing activities, it is crucial to learn the relevance of each feature vector in the sequence when representing activity categories. In this regard, applying attention layers to model the relevant temporal context of activities has proved beneficial in recent HAR studies \cite{attention}. Adopting a similar approach, we utilize a 2-layer \textit{attentional GRU Encoder (AGE)} to process the sequence of refined representations and learn soft attention weights for the generated hidden states $(\mat{h}_t)_{t=1}^{\textrm{T}}$. In the absence of attention mechanism in the temporal domain, classification decision would only be based on the last hidden state achieved after observing the entire sequence. By contrast, empowering the GRU encoder with attention alleviates the burden on the last hidden state and instead, allows learning a holistic summary $\mat{z}$ that takes into account the relative importance of the time-steps 

\begin{equation}
\mat{z} = \sum_{t}\beta_t\mat{h}_{t},
\end{equation} 
where $\beta_t$ denotes the computed attention weight for time-step $t$. Technically, attention values are obtained by first mapping each hidden state into a single score with a linear layer and then normalizing these scores across the time-steps with a softmax function.

\subsection{Center Loss Augmented Objective} \label{sa_sec:centerloss}
Intra-class variability and inter-class similarity are two fundamental challenges of HAR with wearables. The former phenomena occurs since different individuals may execute the same activity differently while the latter challenge arises when different classes of activities reflect very similar sensor patterns. To counter these challenges, the training objective should encourage the model to learn discriminative activity representations; \textit{i.e.}, representations that exhibit large inter-class differences as well as minimized intra-class variations. 

Existing HAR architectures solely rely on the supervision signal provided by the cross-entropy loss during their training phase. While optimizing for this criteria directs the training process towards yielding inter-class separable activity features, it does not explicitly encourage learning intra-class compact representations. To boost the discriminative power of the deep activity features within the learned latent space, we propose to incorporate center-loss \cite{centerloss} for training our HAR model. The auxiliary supervision signal provided by center-loss penalizes the distances between activity representations and their corresponding class centers and thus, reduces intra-class feature variations. Formally, center-loss is defined as 

\begin{equation}
\mathcal{L}_{c}=\frac{1}{2}\sum_{i=1}\|\mat{z}_{i}-\mat{c}_{y_i}\|^2_{2},
\label{sa_eq:loss_c}
\end{equation} 
where $\mat{z}_i\in\mathbb{R}^\mathrm{z}$ denotes the deep representation for sensory segment $x_i$, and $\mat{c}_{y_i}\in\mathbb{R}^\mathrm{z}$ denotes the $y_i$th activity class center computed by averaging the features of the corresponding class. We enforce this criteria on the activity representations obtained from the penultimate layer of our network to effectively pull the deep features towards their class centers. 

In each iteration of the training process, we leverage the joint supervision of cross-entropy loss together with center-loss to simultaneously update the network parameters and the class centers $\mat{c}_{y}$ in an end-to-end manner. Hence, the aggregated optimization objective is formulated as 

\begin{equation}
\Theta^*=\arg\min_{\Theta}\mathcal{L}+\gamma\mathcal{L}_{\textrm{c}},
\label{eq:loss_all}
\end{equation}
where $\mathcal{L}$ represents the cross-entropy loss, $\gamma$ is the balancing coefficient between the two loss functions, and $\Theta$ denotes the collection of all trainable parameters.

\begin{figure*}[tb]
	\centering
	\includegraphics[width=0.6\linewidth]{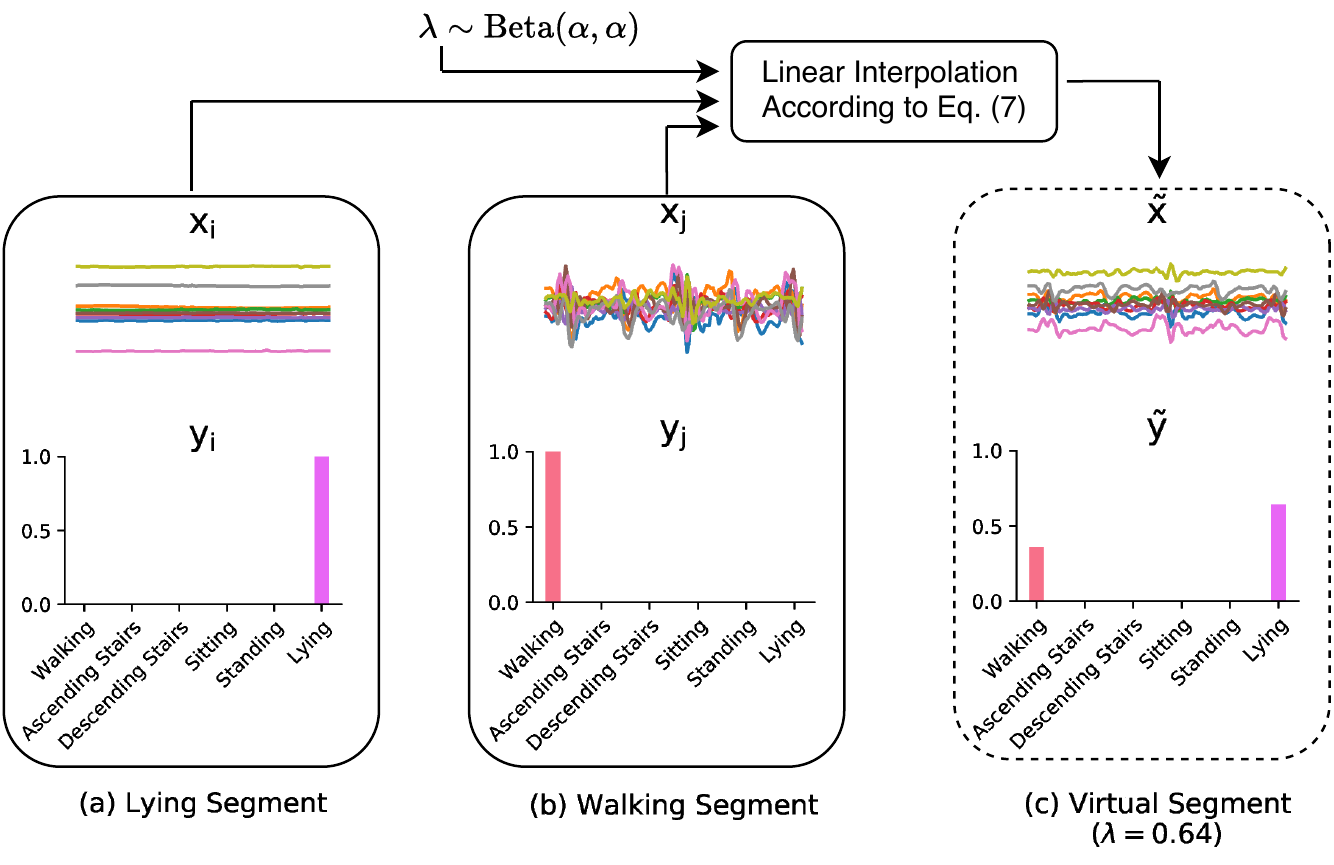}
	\caption{We leverage mixup data augmentation technique to generate virtual sequences during training. We interpolate in-between samples. Here, we visualize (a) a sequence of sensor data from the training split corresponding to the \texttt{lying} activity and its one-hot encoded label representation, (b) a training sensor data segment corresponding to the \texttt{walking} activity, and (c) a \textit{virtual} or generated sequence and its target label according to Eq.~\ref{eq:mixup} with a drawn $\lambda$ value of 0.64 (sampled from a Beta distribution). The visualized data corresponds to a subset of sensor channels in the PAMAP2 dataset~\cite{pamap2}.}

	\label{sa_fig:mixup}
\end{figure*}

\subsection{Mixup Data Augmentation for HAR}\label{sa_sec:mixup}

Due to the laborious task of collecting annotated datasets from wearables, current HAR benchmarks are characterized by their limited sizes. Therefore, introducing new modules and increasing the network parameters without employing effective regularization techniques, makes the model prone to overfitting and endangers its generalization. In this regard, while extending the training data with augmented samples achieved by \textit{e.g.} slight rotations, scaling, and cropping has consistently led to improved generalization performance for computer vision applications, these methods are not applicable to multi-channel time-series data captured by wearables.

In this paper, we explore the effectiveness of a recently proposed data-agnostic augmentation strategy, namely \textit{mixup} \cite{mixup}, for time-series data in order to regularize our deep HAR model. This approach has demonstrated its potential in significantly improving the generalization of deep neural networks by encouraging simple linear behavior in-between training data. In addition, unlike existing augmentation approaches that are dataset dependent and thus require domain expert knowledge for effective adoption, mixup strategy is domain independent and simple to apply. In essence, mixup yields augmented virtual example $(\tilde{x},\tilde{y})$ through linear interpolation of training example pairs $(x_i,y_i)$ and $(x_j,y_j)$, 

\begin{equation}
\begin{aligned}
\tilde{x} &=\lambda x_i + (1-\lambda)x_j \\
\tilde{y} &=\lambda y_i + (1-\lambda)y_j,
\end{aligned}
\label{eq:mixup}
\end{equation}
where $\lambda$ sampled from a Beta($\alpha$, $\alpha$) distribution is the mixing-ratio and $\alpha$ is the mixup hyper-parameter controlling the strength of the interpolation. Notably, mixup augmentation allows efficient generation of virtual examples on-the-fly by randomly picking pairs from the same minibatch in each iteration. In this work, we adopt mixup strategy to augment the time-series segments in each mini-batch and train the model end-to-end with the generated samples. We visually explain the augmentation process with an example in Fig. \ref{sa_fig:mixup}, where a pair of randomly drawn training data sequences are linearly interpolated to yield a novel virtual sequence.
\section{Experiments and Results} \label{sa_sec:experiments}

\begin{figure*}[t]%
	\centering
	\subfloat{\includegraphics[width=0.50\textwidth]{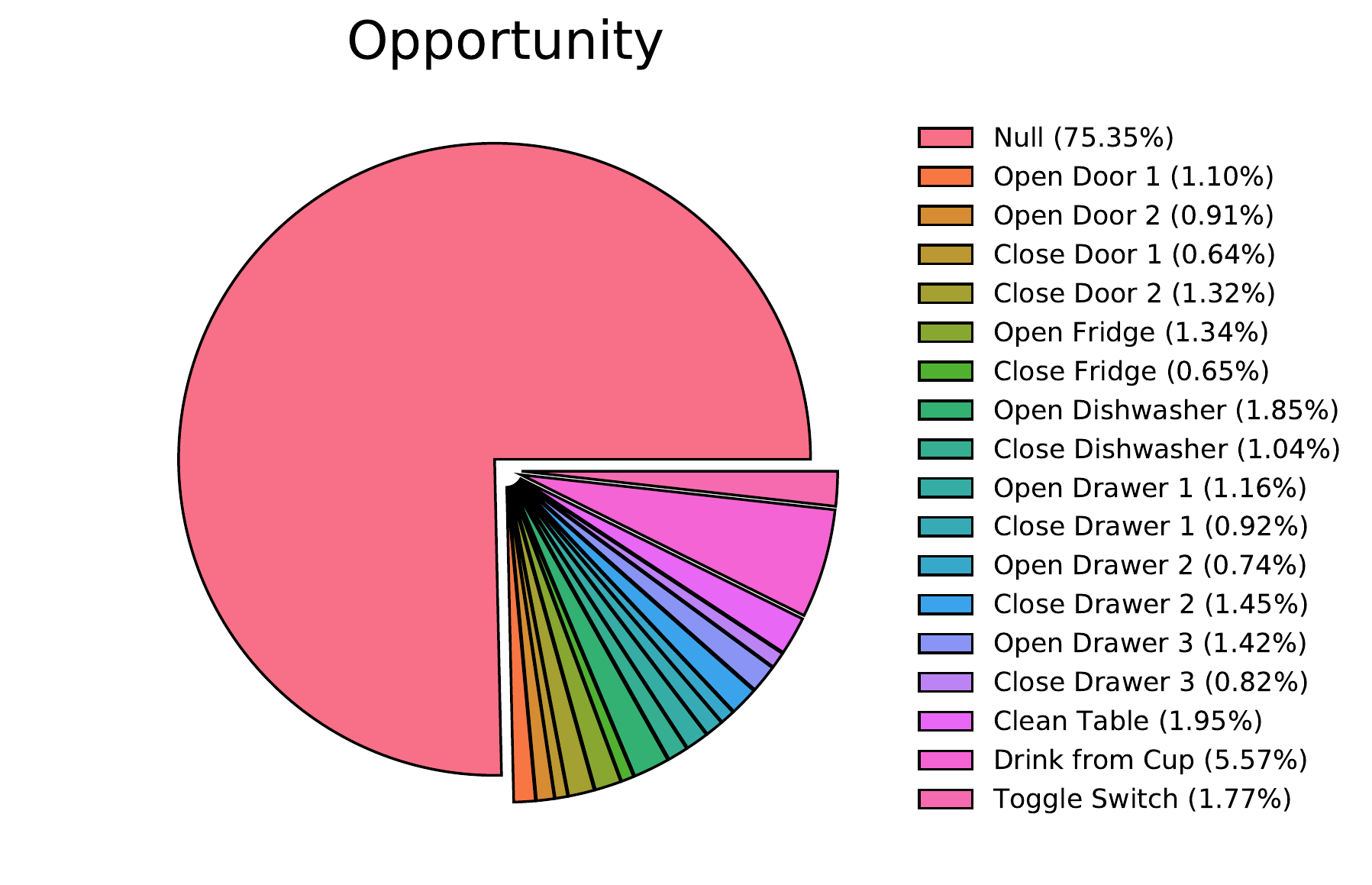}}%
	\subfloat{\includegraphics[width=0.50\textwidth]{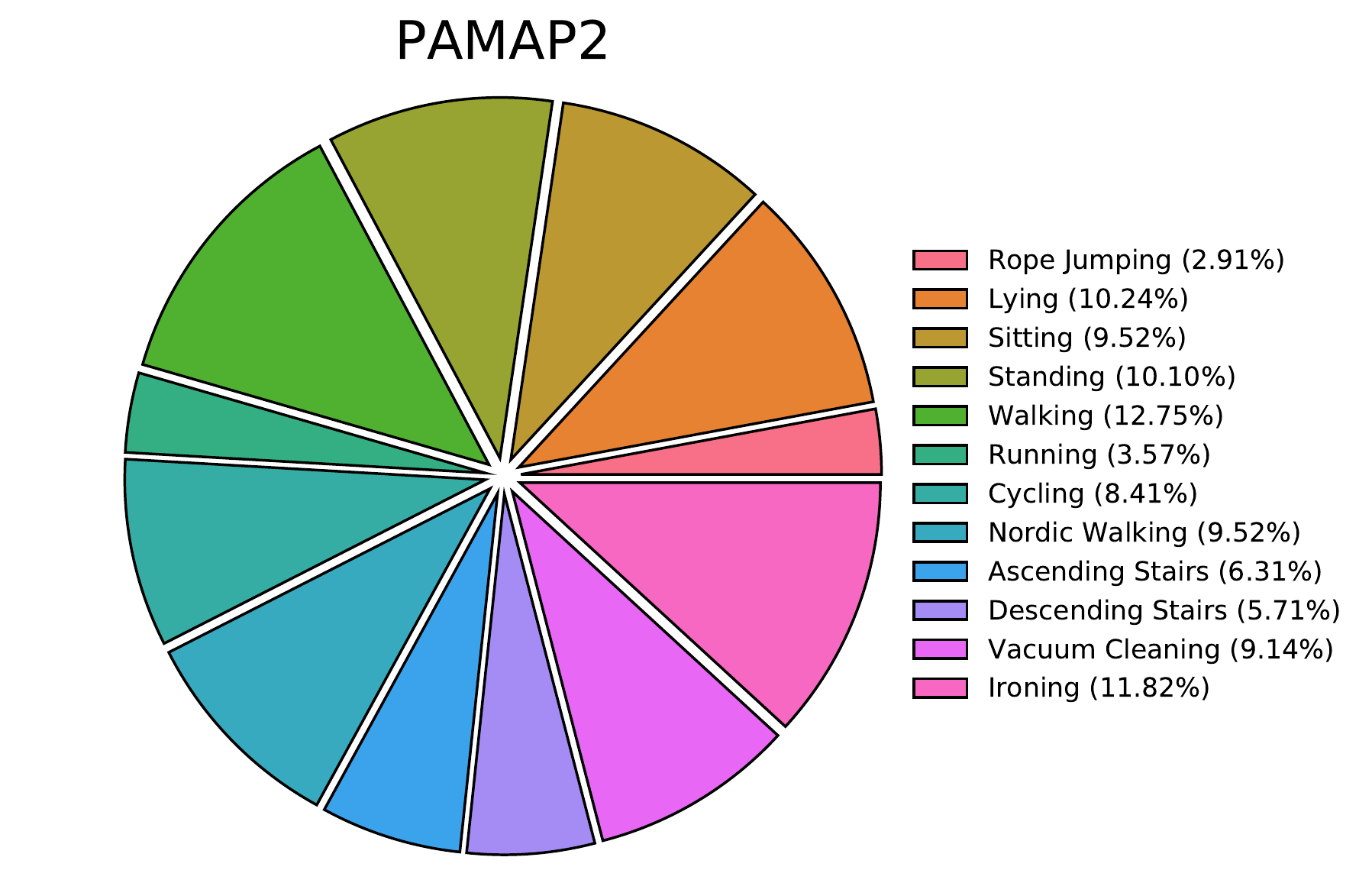}}\\
	\subfloat{\includegraphics[width=0.50\textwidth]{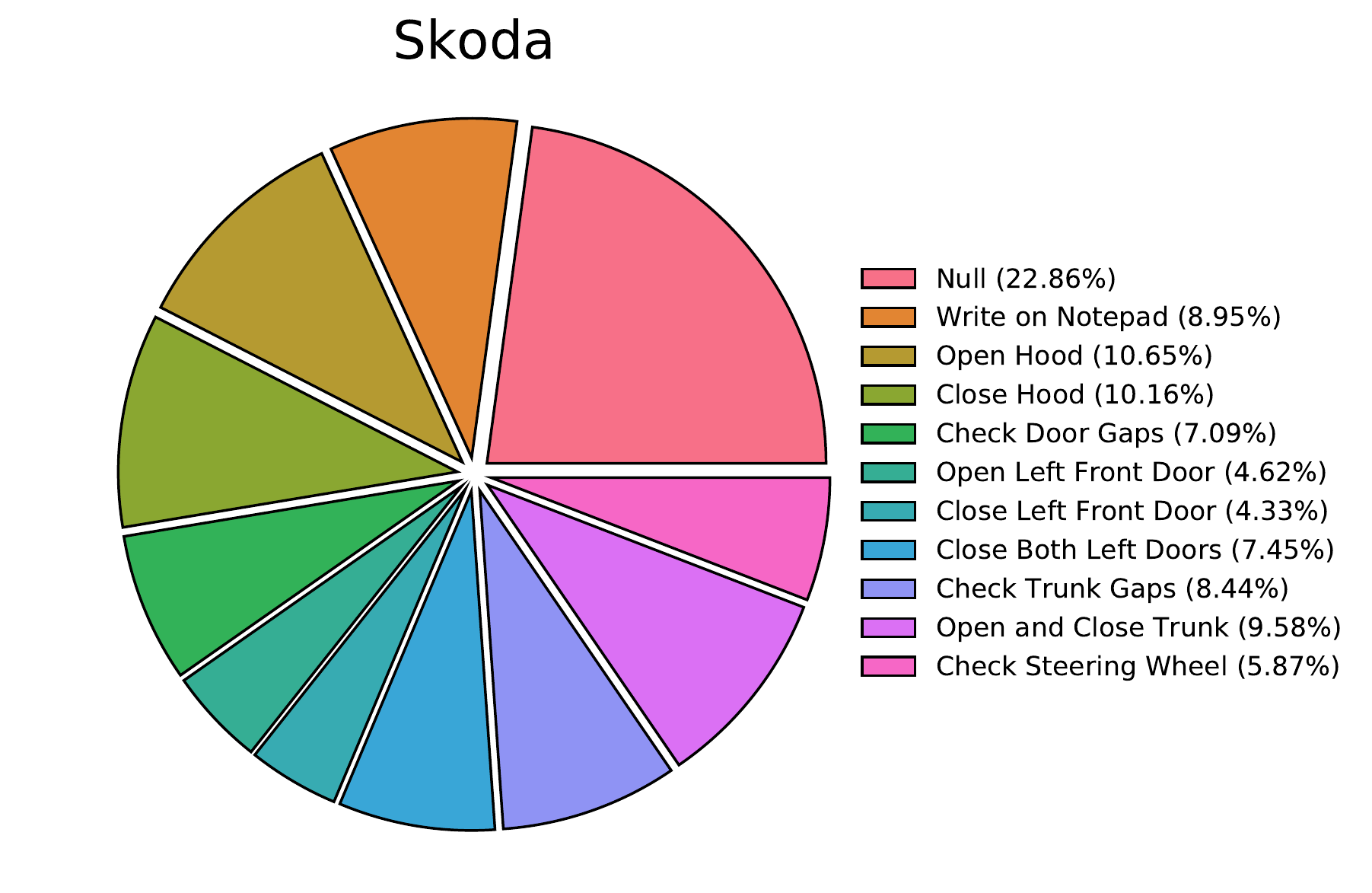}}%
	\subfloat{\includegraphics[width=0.50\textwidth]{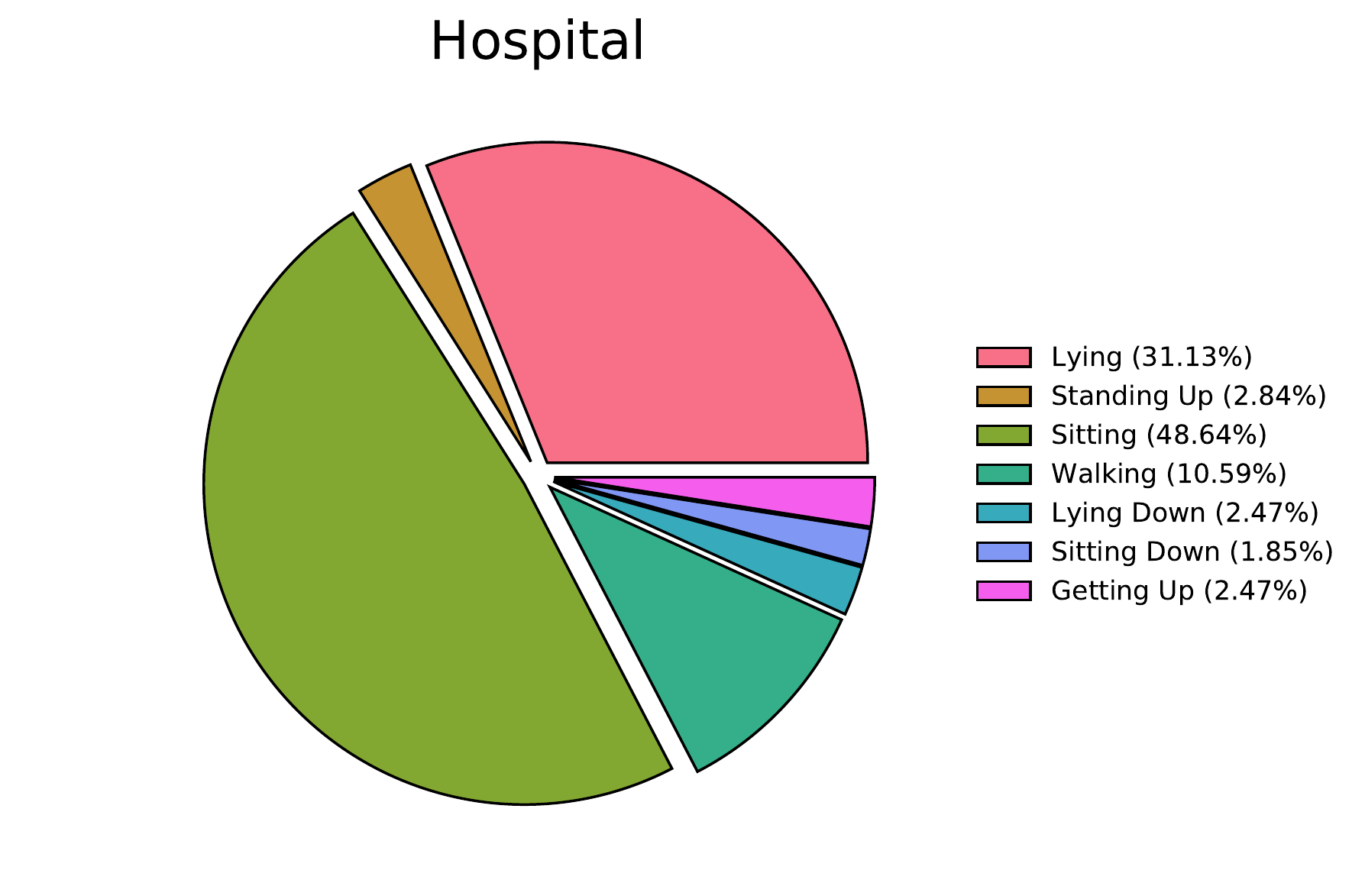}}%
	
	\caption{Benchmark HAR datasets investigated in this paper. We illustrate the activity categories covered and their corresponding distributions within each dataset.}%
	\label{sa_fig:dist}%
\end{figure*}

\subsection{Benchmark Datasets}\label{sa_sub:datasets}
To validate our framework and provide empirical evidence of its generalizability, we employ four HAR benchmarks exhibiting great diversity in terms of the sensing modalities used and the activities to be recognized. We provide a brief description of the datasets in what follows.

\paragraph{Opportunity Dataset \protect\cite{Chavarriaga:2013}.} 
This dataset is captured by multiple body-worn sensors. Four participants wearing the sensors were instructed to carry out naturalistic kitchen routines. The data is recorded at a frequency of 30~Hz and is annotated with 17 sporadic gestures as well as a \texttt{Null} class. Following \cite{bilstm}, the 79 sensor channels not indicating packet-loss are used. For evaluation, we use runs 4 and 5 from subjects 2 and 3 as the holdout test-set, run 2 from participant 1 as the validation-set, and the remaining data as the training-set. 

\paragraph{PAMAP2 Dataset \protect\cite{pamap2}.} This dataset is aimed at recognition 12 diverse activities of daily life. Data was recorded over 52 channels with annotations covering prolonged household and sportive actions. Replicating \cite{bilstm}, we use runs 1 and 2 from subject 6 as the holdout test-set, runs 1 and 2 from subject 5 as the validation-set, and the remaining data for training.

\paragraph{Skoda Dataset \protect\cite{skoda}.} The dataset covers the problem of recognizing $10$ manipulating gestures of assembly-line workers in a manufacturing scenario. Following~\cite{ensemble}, we use the data recorded over 60 sensor channels collected from the right arm, utilize the first $80\%$ of each class for the training-set, the following $10\%$ for validation and the remainder as the test-set.

\paragraph{Hospital Dataset \protect\cite{fcn}.} This dataset is collected from 12 hospitalized older patients wearing an inertial sensor over their garment while performing 7 categories of activities. All the data is recorded at 10~Hz. Following~\cite{fcn}, data from the first eight subjects are used for training, the following three for testing, and the remaining for the validation set. 

We summarize the list of covered activity categories and their corresponding distributions within each dataset in Fig.~\ref{sa_fig:dist}. As illustrated, while the prevalence of each activity category for PAMAP2 and Skoda datasets is quite balanced, we observe a significant distribution imbalance for Opportunity and Hospital datasets.

\subsection{Unified Evaluation Protocol}\label{sa_sec:eval}

To ensure a fair comparison, we directly adopt the evaluation protocol and metrics used in the recent literature~\cite{bilstm,ensemble,attention,role}. Where possible, sensor data are down-sampled to 33~Hz to achieve a consistent temporal resolution with the Opportunity dataset. Each sensor channel is normalized to zero mean and unit variance using the training data statistics. The training data is partitioned into segments using a sliding window of 24 samples (\textit{i.e.}, W=24) with 50\% overlap between adjacent windows. For a realistic setup, sample-wise evaluation is adopted to compare the performance on the test-set; thus, a prediction is made for every sample of the test sequence as opposed to every segment. Given the imbalanced class distributions in the datasets (see Figure~\ref{sa_fig:dist}), the mean F1-score
\begin{equation}
\textrm{F}_{m} = \frac{2}{\mathcal{C}}\sum_{c=1}^{\mathcal{C}}\frac{\textrm{precision}_{c}\times \textrm{recall}_{c}}{\textrm{precision}_{c}+ \textrm{recall}_{c}},
\end{equation} 
is used as the evaluation metric to reflect the ability of the HAR model to recognize every activity category regardless of its prevalence in the collected data. Here, $\mathcal{C}$ denotes the number of activity classes, precision is defined as $\frac{\textrm{TP}}{\textrm{TP}+\textrm{FP}}$, and recall corresponds to $\frac{\textrm{TP}}{\textrm{TP}+\textrm{FN}}$, where $\textrm{TP}$, $\textrm{FP}$, and $\textrm{FN}$ respectively refer to the number of true positives, false positives and false negatives. 

\begin{table}
	\centering
	\caption{A summary of hyper-parameter values selected per dataset. All other hyper-parameters were kept constant across all datasets.}\label{sa_tab:params}
	\resizebox{0.6\columnwidth}{!}{%
		\begin{tabular}{lcccc}
			Hyper-parameter  &     Opportunity & PAMAP2 & Skoda & Hospital \\
			\toprule
			\toprule
			Dropout ratio $p_{\textrm{feat}}$ & 0.5        & 0.9   & 0.5  & 0.5\\
			Dropout ratio $p_{\textrm{cls}}$ & 0.5        & 0.5   & 0.0  & 0.5\\
			Weighting coefficient $\gamma$ & $\textrm{3}\times{\textrm{10}}^{\textrm{-4}}$       & $\textrm{3}\times{\textrm{10}}^{\textrm{-3}}$  & $\textrm{3}\times{\textrm{10}}^{\textrm{-1}}$ & $\textrm{3}\times{\textrm{10}}^{\textrm{-1}}$ \\
			\toprule
		\end{tabular}%
	}
	
\end{table}

\subsection{Implementation Details}
We implement our experiments using Pytorch~\cite{pytorch}. Our network is trained end-to-end for 300 epochs by back-propagating the gradients of the loss function based on mini-batches of size 256 and in accordance with the Adam~\cite{adam} update rule. The learning rate is set to $\textrm{10}^{\textrm{-3}}$ and decayed every 10 epochs by a factor of 0.9. For \textit{mixup} augmentation, we fix $\alpha$=0.8. All these hyper-parameters are kept constant across all datasets. For each dataset, we choose a dropout probability $p\in$\{0, 0.25, 0.5, 0.75, 0.9\} for the refined feature-maps ($p_{\textrm{feat}}$) and the feature vectors fed to the classifier ($p_{\textrm{cls}}$), and select the center-loss weighting coefficient $\gamma\in\textrm{3}\times$\{$\textrm{10}^{\textrm{-4}},\textrm{10}^{\textrm{-3}},\textrm{10}^{\textrm{-2}},\textrm{10}^{\textrm{-1}}$\}, as summarized in Table \ref{sa_tab:params}.

\subsection{Comparisons with the State-of-the-Art}\label{sa_sec:exp_comp}

\begin{table*}[t]
	\centering

	\caption{A comparison of sample-wise activity recognition performance based on class-averaged f1-scores on the holdout test sequences. The baseline results are quoted from \protect\cite{ensemble,attention}, except for (*) where the published code is used with our evaluation protocol.}\label{sa_tab:soa}
	\begin{tabular}{lcccc}
		HAR Model  &     Opportunity & PAMAP2 & Skoda & Hospital* \\
		\toprule \toprule
		LSTM Learner Baseline \cite{ensemble}                 & 65.9        & 75.6   & 90.4  & 62.7 \\
		DeepConvLSTM \cite{deepconvlstm}       & 67.2        & 74.8   & 91.2  & 62.8\\
		b-LSTM-S \cite{bilstm}               & 68.4        & 83.8   & 92.1  & 63.6\\

		Dense Labeling \cite{fcn}*               & 62.4        & 85.4   & 91.6  & 62.9 \\
		Att. Model \cite{attention}              & 70.7        & 87.5   & 91.3  & 64.1 \\
		\toprule
		\textbf{Ours} & \textbf{74.6}        & \textbf{90.8}   & \textbf{92.8} & \textbf{66.6} \\ 
		\textbf{(Improvement over Runner-up)} & \textbf{(5.52\%) }     & \textbf{(3.77\%)}  & \textbf{(0.76\%)}    & \textbf{(3.9\%)} 
	\end{tabular}%

\end{table*}

\begin{figure*}[tb]
	\centering
	\subfloat{\includegraphics[width=0.45\textwidth]{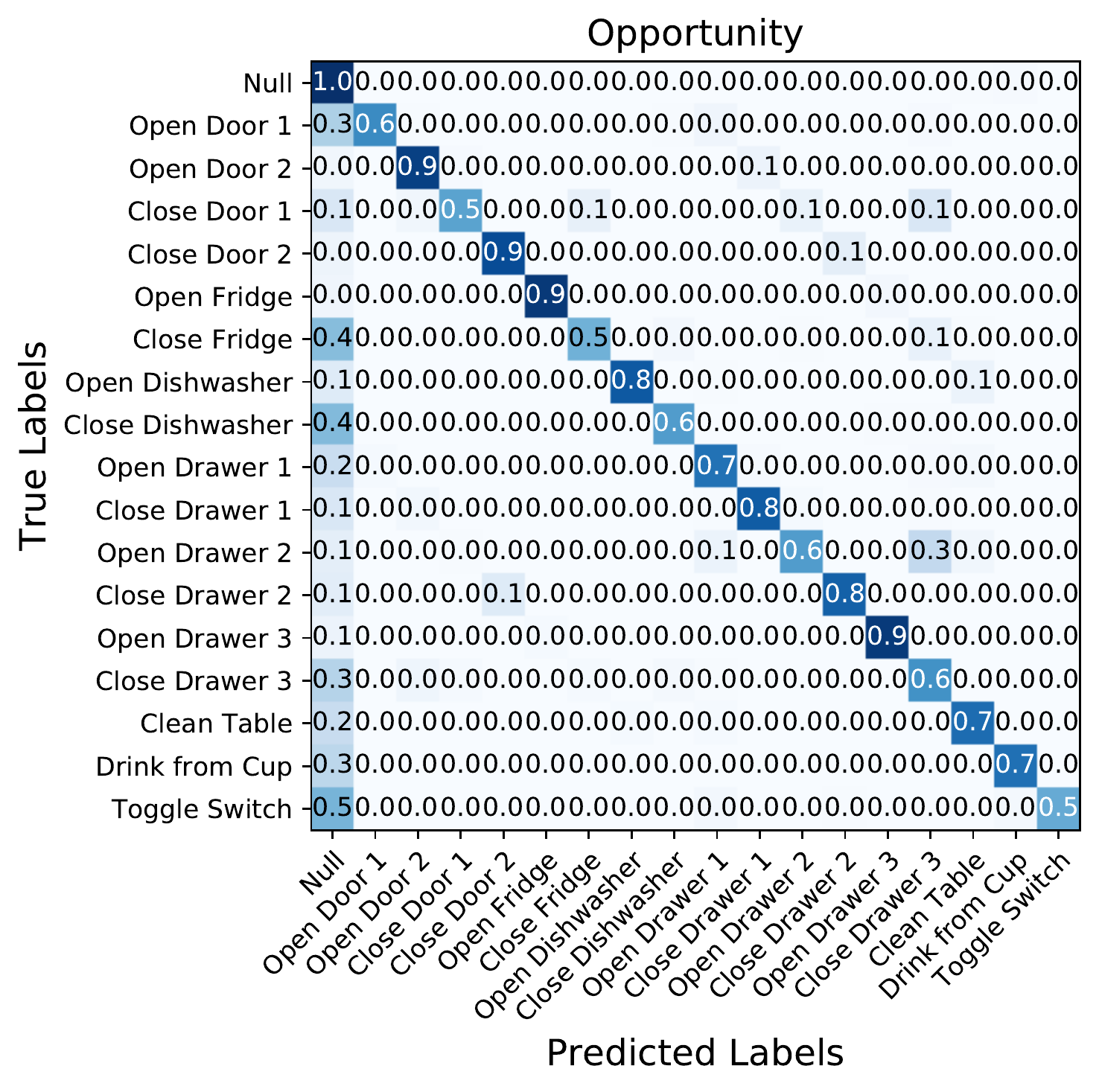}}%
	\subfloat{\includegraphics[width=0.43\textwidth]{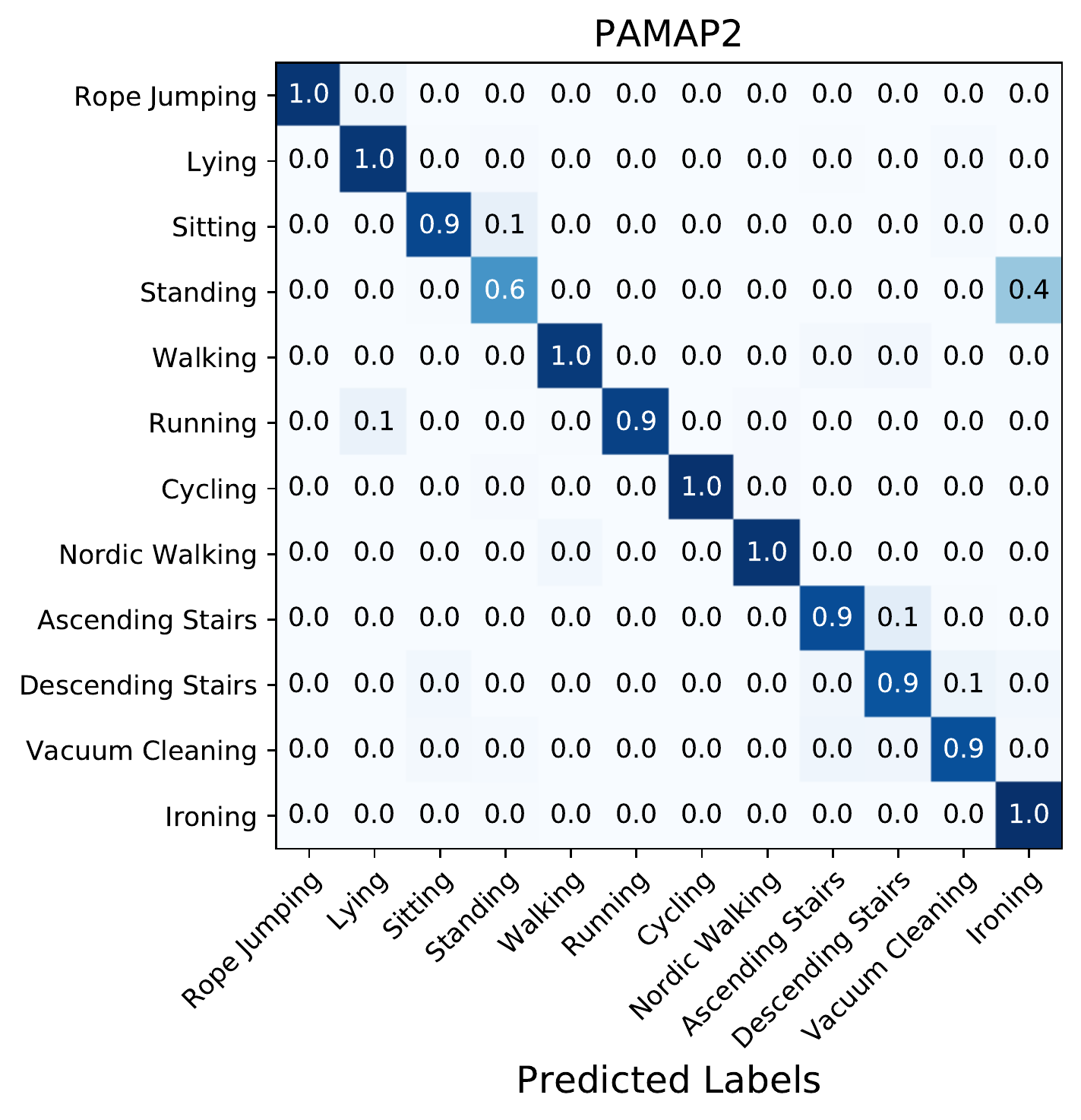}}\\
	\subfloat{\includegraphics[width=0.46\textwidth]{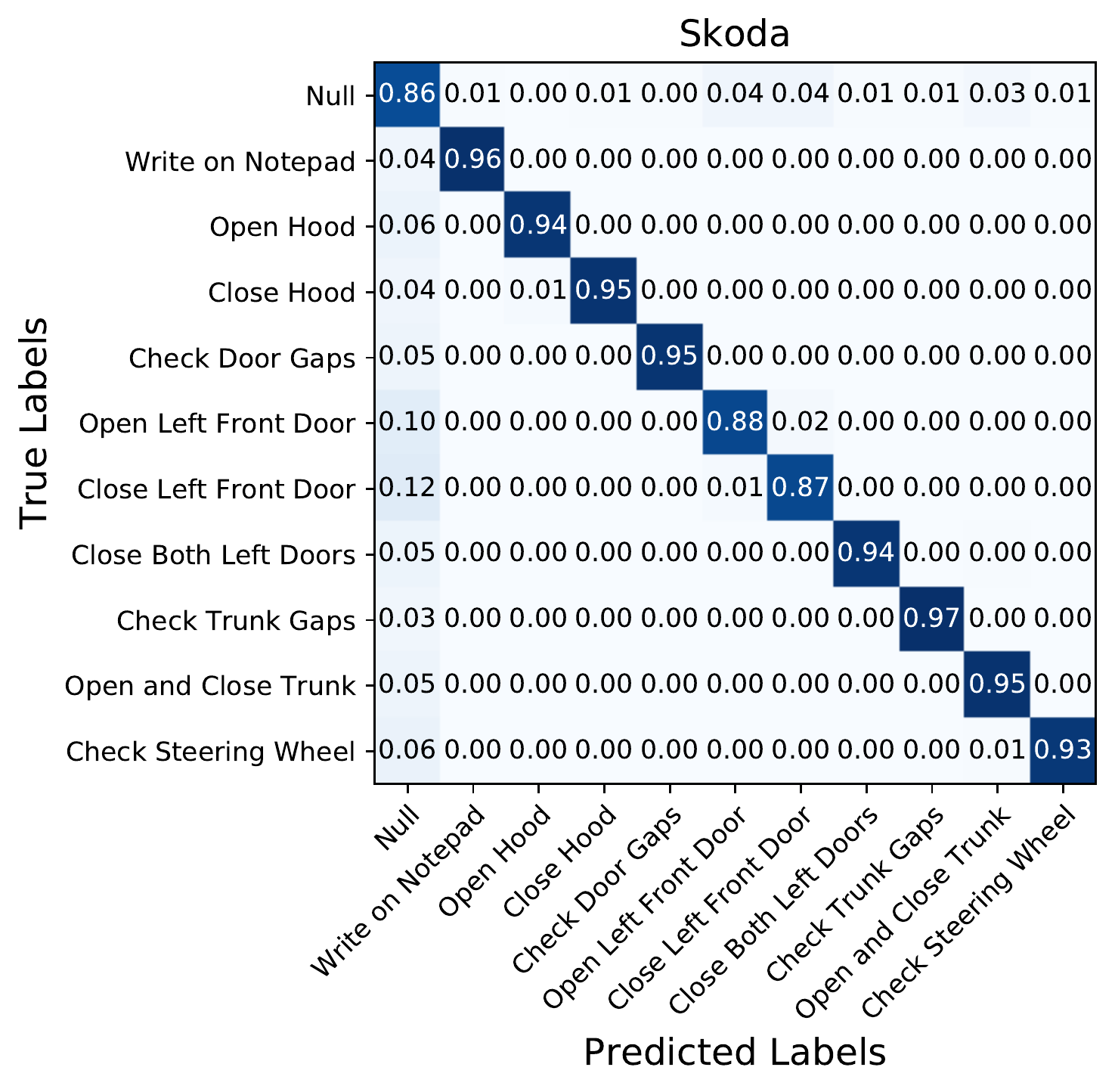}}%
	\subfloat{\includegraphics[width=0.39\textwidth]{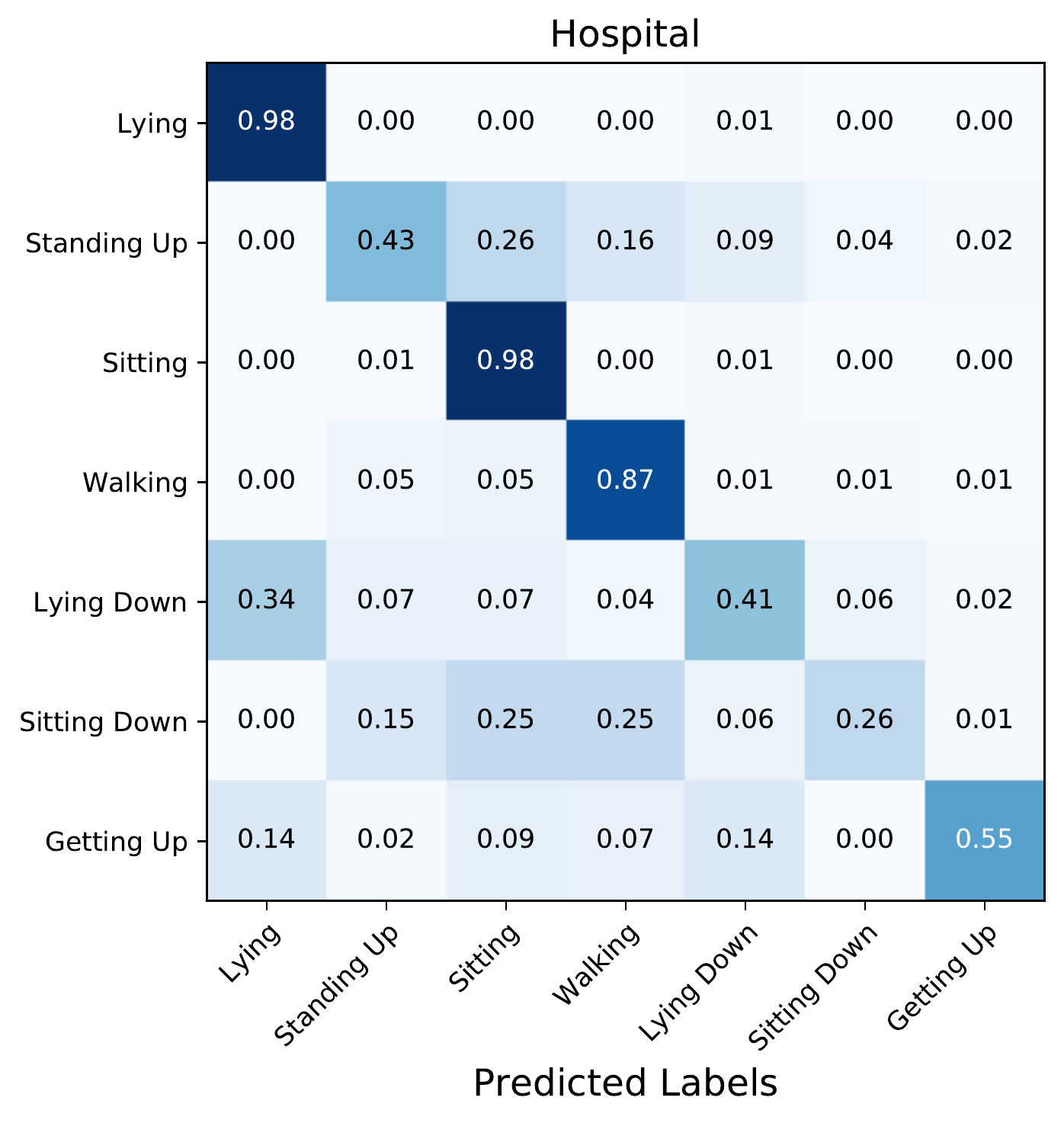}}%
	
	\caption{The confusion matrices highlighting the class-specific recognition performance for the testing splits of Opportunity, PAMAP2, Skoda, and Hospital HAR datasets. The vertical axis represents the ground-truth activity categories and the horizontal axis denotes the predicted activities.}
	\label{sa_fig:cm}%
\end{figure*}

\subsubsection{Classification Measure.}

We compare our proposed framework against state-of-the-art HAR models on four standard benchmarks in Table \ref{sa_tab:soa}. As elucidated in Section \ref{sa_sec:eval}, every baseline generates sample-wise predictions on the entire holdout test sequence and the performance is judged based on the acquired class-averaged f1-score ($\textrm{F}_{m}$). 

In Table \ref{sa_tab:soa}, we can see that the elements we introduced into our framework consistently yield significant recognition improvements over the state-of-the-art models. Interestingly, we observe the highest performance gain of 5.52\% on the Opportunity dataset characterized by $i$) the largest number of incorporated sensor channels; $ii$) the greatest diversity of the actions to recognize; and $iii$) the highest ratio of class imbalance. Our results highlight the significant contribution made  by the integrated components in dealing with challenging activity recognition tasks. Notably, our framework still achieves for a moderate performance improvement on the \textit{performance saturated}~\cite{attention} Skoda dataset. 

For further insights, we summarize the class-specific recognition results from our model by presenting confusion matrices for the four recognition tasks in Fig.~\ref{sa_fig:cm}. We can see that for  Opportunity and Skoda datasets with the inclusion of a \texttt{Null} class in the annotations, most of the confusions occur in distinguishing between the ambiguous \texttt{Null} class and the activities of interest. This can be understood since the \texttt{Null} class represents an infinite number of irrelevant activity data for the HAR problem in hand; thus, explicitly modeling this unknown space is a difficult problem.

\subsubsection{Misalignment Measure}

In addition to the reported classification metrics, we further report on the explicitly designed misalignment measures of \textit{overfill/underfill, insertion,} and \textit{deletion}~\cite{ward} and provide comparisons with the state-of-the-art HAR model~\cite{attention} in Table \ref{sa_tab:misalignment}. These metrics characterize \textit{continuous} activity recognition performance and provide finer details on temporal prediction misalignment with respect to ground truth. Specifically:

\begin{itemize}
    \item \textit{Overfill} and \textit{Underfill} indicate errors when the predicted start or end time of an activity are earlier or later than the ground-truth timings.
    \item \textit{Insertion} errors refer to incorrectly predicting an activity when there is \texttt{Null} activity.
    \item \textit{Deletion} represents wrongly predicting \texttt{Null} class when an activity exists. 
\end{itemize}

Since some measures require the existence of \texttt{Null} class by definition, we report results on Opportunity and Skoda datasets. The quantitative results in Table \ref{sa_tab:misalignment} indicate the improved capability of our model to predict a continuous sequence of activity labels that more accurately aligns with ground-truth timings and better recognizes existence or absence of activities of interest. 

\begin{figure*}[t]%
	\centering
	\subfloat[Opportunity Dataset]{\includegraphics[width=\textwidth]{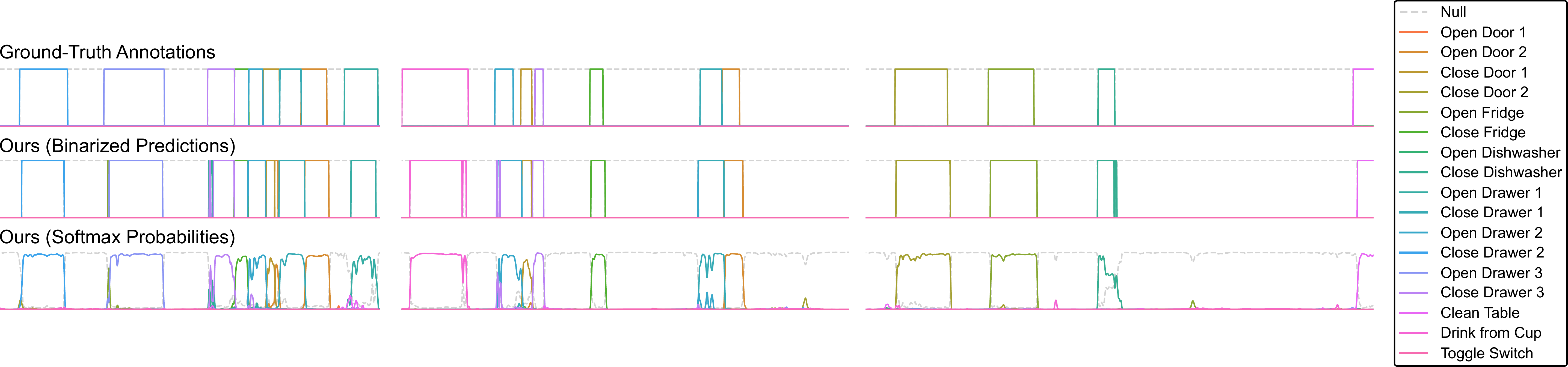}}%
	
	\subfloat[Skoda Dataset]{\includegraphics[width=\textwidth]{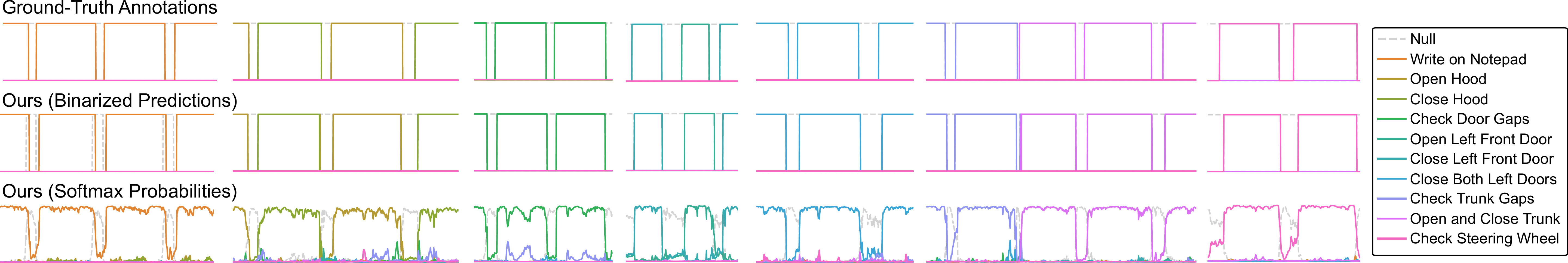}}
	\caption{A visualization of our networks' predictions on the holdout test fragments. Our proposed HAR model accurately localizes and classifies short duration gestures embodied in sequences of sensor signals captured by wearables. We visualize the model's predictions against the ground-truth annotations for sequence fragments of Opportunity and Skoda datasets which include a Null class label representing activities of non-interest.}

	\label{sa_fig:sequence}%
\end{figure*}

Further, we visualize fragments of sensor recordings from these datasets in Fig. \ref{sa_fig:sequence} for qualitative assessment. The Skoda dataset includes repetitive execution of quality check gestures while the Opportunity dataset is characterized by short duration and sporadic activities. We present the ground-truth annotations (top rows), our model's softmax output probabilities (last rows) and the binarized sequence of predictions (middle rows) obtained after applying argmax operation on the soft scores for each time-step. At every time-step, we color-code and plot the output class probabilities for each activity category, where we observe a strong correspondence between the ground-truth annotations, activity duration and the predicted activity scores. 

\begin{table}
	\centering
	
	\caption{Misalignment measures comparison. (*) denotes the best performing state-of-the-art recognition model \protect\cite{attention} according to Table \ref{sa_tab:soa}.}
	\label{sa_tab:misalignment}%
	\resizebox{0.5\textwidth}{!}{%
	\begin{tabular}{lcc|cc}
		
		& \multicolumn{2}{c}{Opportunity} &  \multicolumn{2}{c}{Skoda} \\
		Alignment Measures                          & Ours        & SoA*      &  Ours     & SoA*     \\
		\toprule
		\toprule
		Deletion  ($\downarrow$)      &\textbf{0.62} & 0.69   &\textbf{0.04} &\textbf{0.04}                   \\
		Insertion  ($\downarrow$)     &\textbf{2.87} & 3.34     &\textbf{2.01} & 3.34                \\
		Underfill/Overfill ($\downarrow$) &\textbf{3.71} & 4.15              &5.33 & \textbf{5.17}            \\
		True Positives ($\uparrow$)   &\textbf{92.8} & 91.82               &\textbf{92.62} &91.45                \\
		
	\end{tabular}%
	}

\end{table}

\begin{table}
	\centering
	\caption{We investigate the contribution of integrated modules by conducting an ablation study on the Opportunity dataset.}\label{sa_tab:ablation}
	\resizebox{0.5\textwidth}{!}{%
	\begin{tabular}{lc}
		HAR Model  & $\textrm{F}_{m}$ \\
		\toprule \toprule
		DeepConvLSTM Baseline                 
		& 67.2    \\
		Ours (mixup)                 
		& 70.7     \\
		
		Ours (mixup + CenterLoss)
		& 72.2     \\
		Ours (mixup + AGE)
		& 71.7     \\
		
		Ours (mixup + CIE)              
		& 73.0     \\
		
		Ours (mixup + CenterLoss + AGE)               
		& 72.3     \\
		
		Ours (mixup + CenterLoss + CIE )  
		& 73.2     \\
		
		Ours (mixup + CIE + AGE)              
		& 74.0     \\
		\toprule
		
		\textbf{Ours (mixup + CenterLoss + CIE + AGE ) } & \textbf{74.6}    
	\end{tabular}%
	}

\end{table}

\subsection{Ablation Studies and Insights} \label{sa_sec:ablation}

Given that our proposed HAR model integrates several key ideas into a single framework, we conduct an ablation study on the Opportunity dataset to understand the contribution made by the various components for the human activity recognition task in Table~\ref{sa_tab:ablation}. For each ablated experiment, we remove specific modules of our framework and as a reference we include DeepConvLSTM---the backbone of our network as illustrated in Fig.~\ref{sa_fig:arch}. 

Unsurprisingly, removing any component handicaps the HAR model and reduces performance (to 67.2\%---see DeepConvLSTM baseline performance) while incorporating all components together yields the highest performing HAR model (74.6\%---see \textsf{mixup+CenterLoss+CIE+AGE}). 
Notably, the effectiveness of time-series data augmentation in regularizing HAR models can be realized as \textsf{mixup} alone results in a significant relative improvement of 5.2\% over the DeepConvLSTM Baseline (from 67.2\% to 70.7\%). The virtual multi-channel time-series data attained through in-between sample linear interpolations expand the training data and effectively improve the generalization of learned activity features to unseen test sequences. 

\begin{figure}[t]
	\centering
	\includegraphics[width=0.9\columnwidth]{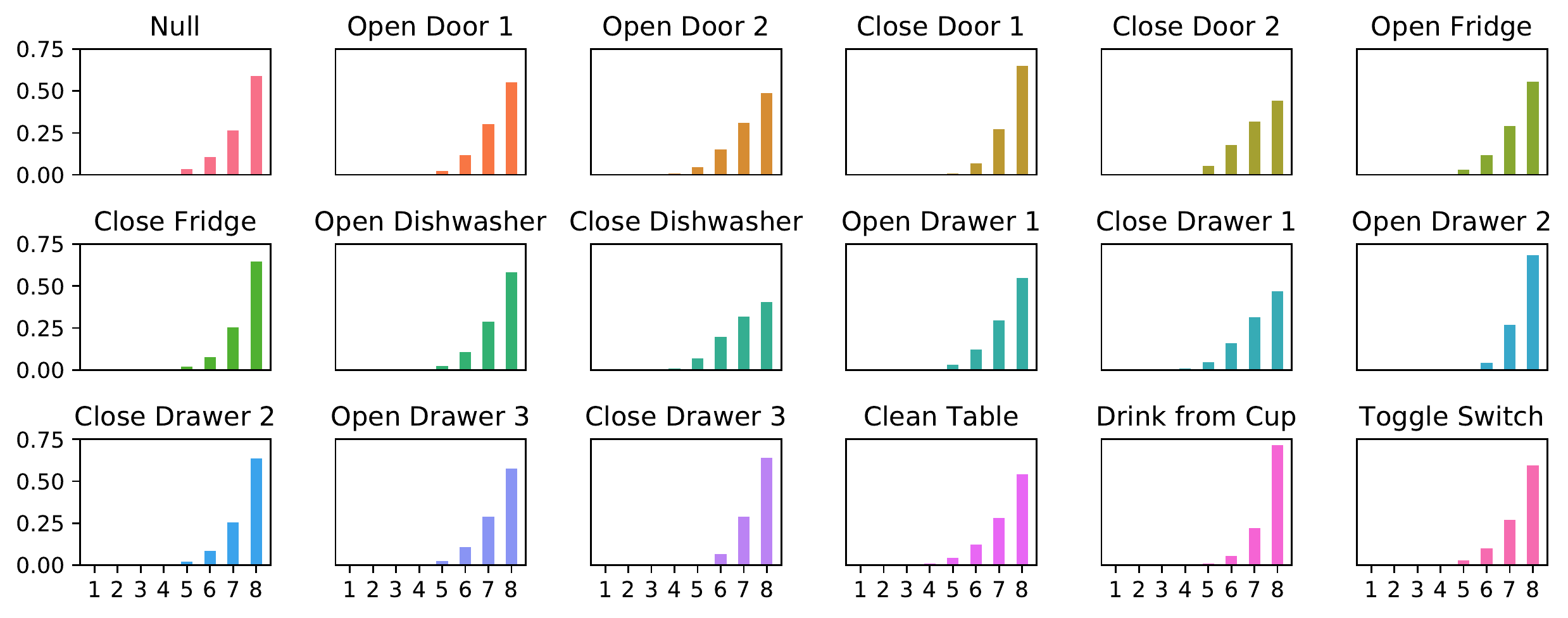}
	\caption{A visualization of discovered temporal attentions by our AGE module. The vertical axis represents the normalized attention scores and the horizontal axis denotes hidden states $(\mat{h}_t)_{t=1}^{\textrm{T}=8}$ of our GRU encoder.}

	\label{sa_fig:ta}
\end{figure}

As hypothesized, encouraging minimal intra-class variability of representations with \textit{center-loss} consistently improves the recognition performance for activities (\textsf{mixup+CenterLoss}). In addition, while both \textit{CIE} and \textit{AGE} modules allow learning better representations of activities reflected by the enhanced metrics for \textsf{mixup+CIE+AGE} compared to \textsf{mixup} (4.7\%  relative improvement), we observe a larger performance gain when incorporating \textit{CIE} module as compared with \textit{AGE}; the former encodes the cross-channel sensor interactions with self-attention while the latter learns the relevance of time-steps with temporal attention. Presumably this is due to the fact that within the Opportunity challenge setup, the sequence of representations fed to the GRU is quite short in length (\textit{i.e.}, T=8) and therefore, the last hidden state alone captures most of the information relevant to the activity. In order to verify this, we extract the learned attention scores $\beta_t$ corresponding to the hidden states $(\mat{h}_t)_{t=1}^{\textrm{T}=8}$ of our GRU encoder and present an illustration for every activity category of Opportunity dataset in Fig.~\ref{sa_fig:ta}. In line with the observations made in \cite{attention}, the recurrent neural network progressively becomes more informed about the activity and thus, proportionally places higher attention on the few last hidden states with the last state dominating the attendance.

\begin{figure}[t]
	\centering
	\includegraphics[width=0.8\columnwidth]{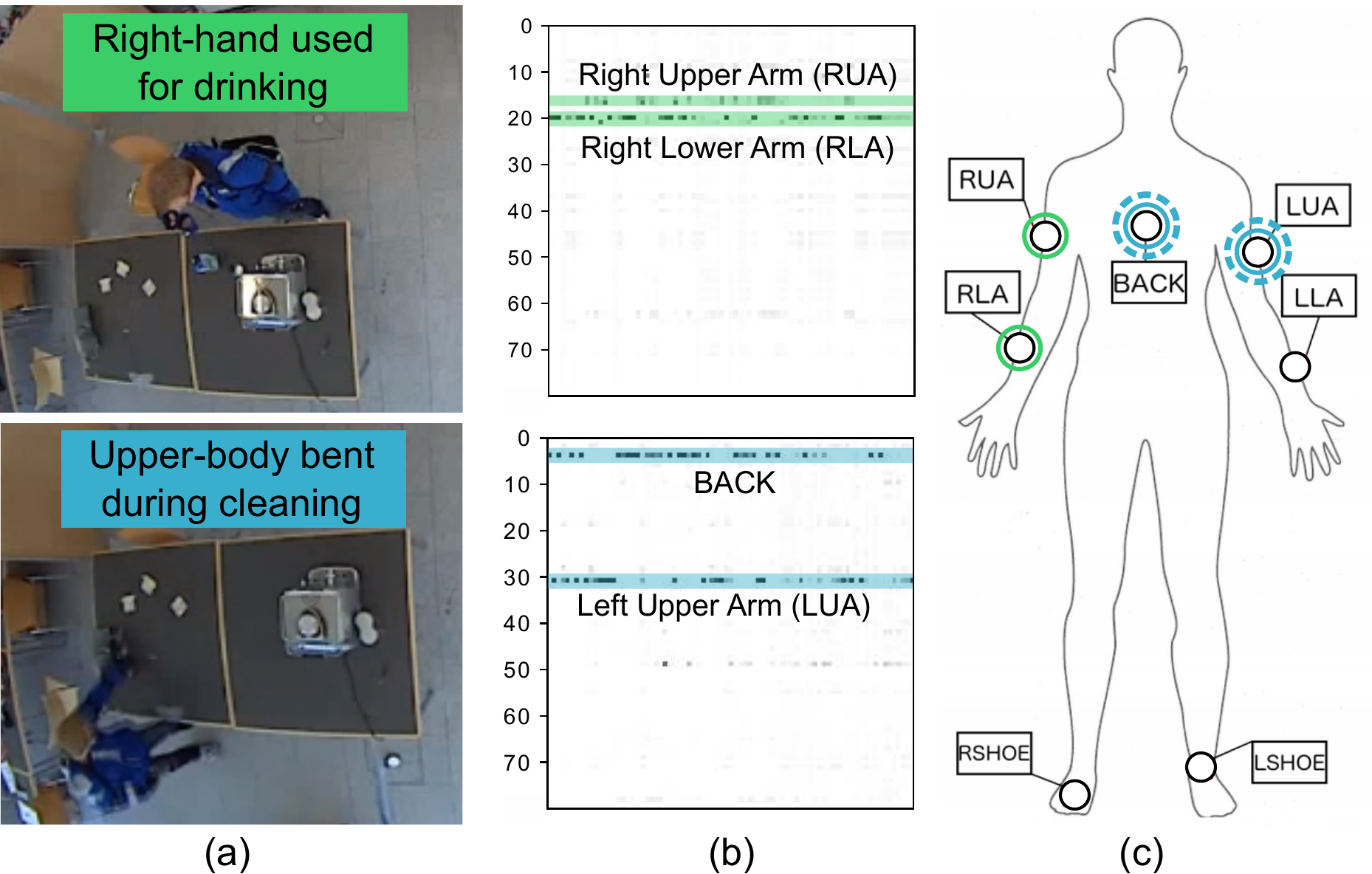}
	\caption{A visualization of learned self-attention correlations by our CIE module. (a) Subject engaged in two activities, (b) discovered cross-channel correlations by our model selected for \textit{Right-hand used for} \texttt{drinking from cup} (dark shaded marks along the rows highlighted in green) and \textit{Upper-body bent during} \texttt{cleaning table} (dark shaded marks along the rows highlighted in blue) as shown in video snapshots recorded during the data collection process and shown in (a), and (c) highly attended sensor locations for each activity (color-coded) in the Opportunity dataset.}

	\label{sa_fig:sa}
\end{figure}

On the other hand, we observed exploiting latent channel interactions to significantly improve activity representations as highlighted in ablation results in Table~\ref{sa_tab:ablation}. To visually explain the learned self-attention correlations from our proposed cross-channel encoder, we graph two segments associated with activity classes of \texttt{drinking from cup} and \texttt{cleaning table}. The CIE module consumes an input sequence and generates a normalized score matrix of size D$\times$D, corresponding to the attention between each pair of D=79 channels. In Fig.~\ref{sa_fig:sa}, we present the normalized self-attention scores, $\mat{a}\in\mathbb{R}^{\textrm{79}\times\textrm{79}}$ (attained from softmax operation) in Eq. \ref{sa_eq:att}, where each column in the attention matrix indicates the extent that a particular sensor channel attends to available sensor channels. 

We observe a clear and meaningful focus on a subset of channels vital to the recognition of activities indicated by dark rows in the matrices. For example, we notice high attendance: \textit{i}) to the inertial measurement units (IMUs) on the right arm when \textit{right hand is being used for} \texttt{drinking from cup}; and \textit{ii}) to the IMUs placed on the back and left-upper arm when \textit{upper-body is bent during} \texttt{cleaning table}. Thus, the explicit modeling of sensor channel interactions not only leads to improved recognition performance as substantiated by our ablation study in Table~\ref{sa_tab:ablation}, but also facilitates visual explanation through interpretable scores. 
\section{Conclusions}
Human activity recognition (HAR) using wearables has created increasingly new opportunities for healthcare applications. We present a new HAR framework and demonstrate its effectiveness through significant performance improvements achieved over state-of-the-art and its generalizability by evaluations across four diverse benchmarks. In particular, we: $i$) enriched activity representations by exploiting latent correlations between sensor channels; $ii$) incorporated center-loss to alleviate dealing with intra-class variations of activities; and $iii$) augmented multi-channel time-series data with mixup for better generalization. 
We believe that our work will provide new opportunities to further research in HAR using wearables.

\bibliographystyle{unsrt}
\bibliography{main}

\end{document}